\documentclass[letterpaper, 10 pt, conference]{ieeeconf}  

\IEEEoverridecommandlockouts                              

\usepackage{makeidx}         
\usepackage{graphicx}        
\usepackage{multicol}        

\usepackage{amssymb}
\usepackage{amsmath}
\DeclareMathOperator*{\argminA}{arg\,min}

\newcommand{\spa}{{sense-plan-act}}
\newcommand{\SPA}{{Sense-Plan-Act}}
\newcommand{\Spa}{{Sense-plan-act}}

\newcommand{\rp}{{reactive planning}}
\newcommand{\RP}{{Reactive Planning}}
\newcommand{\Rp}{{Reactive planning}}

\newcommand{\lrc}{{locally reactive control}}
\newcommand{\LRC}{{Locally Reactive Control}}
\newcommand{\Lrc}{{Locally reactive control}}

\newcommand{\CMO}{{Continuous Motion Optimization}}

\usepackage{moreverb,url}

\usepackage[colorlinks,bookmarksopen,bookmarksnumbered,citecolor=red,urlcolor=red]{hyperref}

\newcommand\BibTeX{{\rmfamily B\kern-.05em \textsc{i\kern-.025em b}\kern-.08em
T\kern-.1667em\lower.7ex\hbox{E}\kern-.125emX}}

\usepackage{amsmath}

\usepackage{xcolor}
\usepackage{tikz}
\usetikzlibrary{bayesnet}
\usetikzlibrary{arrows}

\usepackage{subcaption}
\definecolor{lrc_color}{HTML}{AA0000}
\definecolor{rp_color}{HTML}{FFAA00}
\definecolor{perception_color}{HTML}{1542BE}
\definecolor{cmo_color}{HTML}{00975A}

\title{\huge{Real-time Perception meets Reactive Motion Generation}}
\author{Daniel Kappler$^{1,3}$, Franziska Meier$^{1,2,5}$, Jan Issac$^{1,3}$, Jim Mainprice$^{1,4}$,
  Cristina Garcia Cifuentes$^{1}$, \\ Manuel W{\"u}thrich$^{1}$, Vincent Berenz$^{1}$,
  Stefan Schaal$^{1,2}$, Nathan Ratliff$^{3}$ and Jeannette Bohg$^{1,6}$
\thanks{$^{1}$ Autonomous Motion Dept. at the MPI for Intelligent Systems, Germany Email: {\tt\small   first.lastname@tue.mpg.de}}
\thanks{$^{2}$ CLMC lab at the Univ. of Southern California, USA}
\thanks{$^{3}$~Lula Robotics Inc., USA}%
\thanks{$^{4}$~Univ. of Stuttgart, Germany}%
\thanks{$^{5}$~Dept. of Computer Science \& Engineering,
  Univ. of Washington, USA}%
\thanks{$^{6}$~Dept. of Computer Science, Stanford Univ., USA}}

\begin{document}

\maketitle

\begin{abstract}
  We address the challenging problem of robotic grasping and
  manipulation in the presence of uncertainty. This uncertainty is due
  to noisy sensing, inaccurate models and hard-to-predict environment
  dynamics. We  quantify the importance of continuous,
  real-time perception and its tight integration with reactive motion
  generation methods in dynamic manipulation scenarios. We compare
  three different systems that are instantiations of the most common
  architectures in the field: (i) a traditional {\em sense-plan-act\/}
  approach that is still widely used, (ii) a myopic controller that
  only reacts to local environment dynamics and (iii) a reactive
  planner that integrates feedback control and motion optimization.
  All architectures rely on the same components for real-time
  perception and reactive motion generation to allow a
  quantitative evaluation. We extensively evaluate the systems on a real
  robotic platform in four scenarios that exhibit either a challenging
  workspace geometry or a dynamic environment. In 333 experiments, we
  quantify the robustness and accuracy that is due to integrating
  real-time feedback at different time scales in a reactive motion
  generation system. We also report on the lessons learned for system
  building. 
\end{abstract}

\section{Introduction}
Robotic systems that integrate perceptual feedback into their
planning and control loops have been developed for relatively
low-dimensional problems such as autonomous driving or flying
\cite{leonard2008perception,Richter2016}. These systems are now mature 
enough to be on the verge of becoming consumer products.  
For problems that require controlling many degrees of freedom (DoF)
and physically interacting with the environment, it remains an open
question {\em how\/} to best integrate perception and motion
generation to allow for reactive behavior in the face of
uncertainty. This is despite the fact that high-performance components
for both visual tracking and reactive planning have been 
proposed in recent years. The teams who participated in the recent
robotics challenges ({\em DARPA Robotics Challenge\/}
(DRC) \cite{drc_chris,drc_ihmc} or {\em Amazon Picking
  Challenge\/} (APC) \cite{apc_2016,APC_AllTeams}) testified to this
insight. 

In this paper, we present an instantiation of a robotic system which
tightly integrates real-time\footnote{As common in the Computer Vision
  area, we use the term {\em real-time\/} to indicate that the
  computation time required by the perception methods are below the
  frame-rate of the depth camera, i.e. below 30Hz.} perception with
reactive motion generation for autonomous manipulation.
We use visual perception to simultaneously track the target object and
robot arm, and to obtain a geometrical representation of the workspace
obstacles.
The object pose, workspace geometry and robot configuration are then
consumed by both, a local controller at a high rate (1 KHz) and a
continuous motion optimizer at a lower rate (5-10 Hz).

{\em This is a systems paper.\/} It aims at drawing conclusions on
system integration based on empirical evidence from an extensive 
number of experiments. 
We quantify the benefit of integrating real-time perceptual feedback
and reactive motion generation in dynamic manipulation scenarios for
high DoF systems.
We compare to baseline systems that rely on the same
components but process sensory information at different rates or
optimize the motion over different time horizons.
This paper proposes requirements for components rather than prescribing
specific perceptual, planning or control modules. 
This quantitative evaluation on the level of integration is one aspect that
makes this paper unique in relation to related work on robotic manipulation
systems.  

Another aspect is the complexity in our experimental
scenarios: they contain dynamic target objects and a
dynamic environment - conditions that are common in human-robot
collaboration, the household, or 
disaster relief scenarios. Compared to the aforementioned robot challenges
(APC and DRC), our experimental scenarios consider a much smaller
variety of manipulation tasks and are tested on a fixed-base platform.
However in terms of dynamicity, our scenarios go beyond those
considered in the challenges where the environment is static and only
the robot interacts with it. Furthermore, we do not use any
teleoperation as was the case in the DRC.
We extensively evaluate the
adaptivity, accuracy and robustness of the alternative systems in
four different scenarios with varying degrees of geometric complexity and
dynamicity. We draw the following conclusions:
(i) Incorporating real-time feedback on different time-scales is crucial to
  achieve safe and successful task execution in uncertain, dynamically changing
  environments, (ii) the availability of reactive motion
  generation relaxes the requirement for perception systems to achieve
  maximum, one-shot accuracy since new, updated information can be consumed
  immediately.

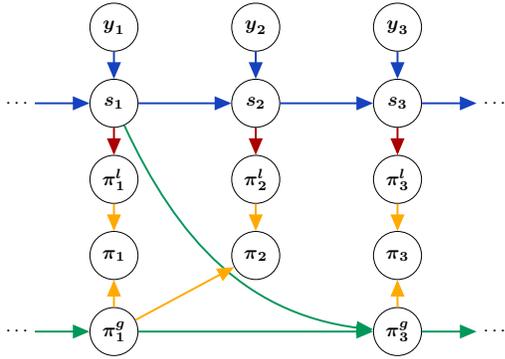
\begin{figure}[!t]
  \centering
    \resizebox{0.8\columnwidth}{!}{\newcommand{\one}{\boldsymbol{1}}%
\newcommand{\two}{\boldsymbol{2}}%
\newcommand{\three}{\boldsymbol{3}}%
\newcommand{\obs}{ \boldsymbol{y }}%
\newcommand{\sta}{ \boldsymbol{s }}%
\newcommand{\con}{ \boldsymbol{\pi }}%
\newcommand{\loc}{ \boldsymbol{\pi}^{\boldsymbol{l}} }%
\newcommand{\glo}{ \boldsymbol{\pi}^{\boldsymbol{g}} }%

\tikzstyle{node_style} = [circle,fill=white,draw=black,inner sep=1pt,
minimum size=25pt,  node distance=1.7]%

\tikzstyle{perception} = [perception_color, line width = 1pt]%
\tikzstyle{local} = [lrc_color, line width = 1pt]%
\tikzstyle{global} = [cmo_color, line width = 1pt]%
\tikzstyle{fusion} = [rp_color, line width = 1pt]%

\begin{tikzpicture}
   \node[node_style]                   	  	(obs_1) {${\obs_{\one}}$};
   \node[node_style, right=of obs_1]          		(obs_2) {${\obs_{\two}}$};
   \node[node_style, right=of obs_2]          		(obs_3) {${\obs_{\three}}$};

   \node[node_style, below=0.5cm of obs_1]             	(sta_1) {${\sta_{\one}}$};
   \node[node_style, below=0.5cm of obs_2]	          	(sta_2) {${\sta_{\two}}$};
   \node[node_style, below=0.5cm of obs_3]   	       	(sta_3) {${\sta_{\three}}$};
    
   \node[node_style, below=0.5cm of sta_1]                  		(loc_1) {${\loc_{\one}}$};
   \node[node_style, below=0.5cm of sta_2, right=of loc_1]         (loc_2) {${\loc_{\two}}$};
   \node[node_style, below=0.5cm of sta_3, right=of loc_2]         (loc_3) {${\loc_{\three}}$};
   
   \node[node_style, below=0.5cm of loc_1]                      	(con_1) {${\con_{\one}}$};
   \node[node_style, below=0.5cm of loc_2, right=of con_1]   	(con_2) {${\con_{\two}}$};
   \node[node_style, below=0.5cm of loc_3, right=of con_2]          (con_3) {${\con_{\three}}$};
   

   \node[node_style, below=0.5cm of con_1]                      	(glo_1) {${\glo_{\one}}$};
   \node[node_style, below=0.5cm of con_3]          (glo_3) {${\glo_{\three}}$};

   \node[left=of sta_1](sta_past) {${\cdots}$};
   \node[right=of sta_3](sta_future) {${\cdots}$};

   \node[left=of glo_1](glo_past) {${\cdots}$};
   \node[right=of glo_3](glo_future) {${\cdots}$};

  \edge[perception] {sta_1} {sta_2};
  \edge[perception] {sta_2} {sta_3};
  \edge[perception] {sta_past} {sta_1.west} ; %
  \edge[perception] {sta_3.east} {sta_future}; %
  
  \edge[perception] {obs_1} {sta_1};
  \edge[perception] {obs_2} {sta_2};
  \edge[perception] {obs_3} {sta_3} ; %
  
  \edge[local]{sta_1} {loc_1} ;
  \edge[local]{sta_2} {loc_2} ;
  \edge[local]{sta_3} {loc_3}  ; %

  \edge[global] {glo_1} {glo_3};
  \edge[global,out=-65,in=175] {sta_1} {glo_3};
  \edge[global] {glo_past} {glo_1.west} ; %
  \edge[global] {glo_3.east} {glo_future}; %
  
  \edge[fusion] {loc_1}{con_1}
  \edge[fusion] {glo_1}{con_1}
  \edge[fusion] {loc_2}{con_2}
  \edge[fusion] {glo_1}{con_2}
  \edge[fusion] {loc_3}{con_3}
  \edge[fusion] {glo_3}{con_3}


  \coordinate (stretcher) at (0,-1.1);
\end{tikzpicture}

}
  \caption{Flow of information across three time steps: 
  The \textcolor{perception_color}{perception modules} continuously
  infer the state of the robot and the world $\boldsymbol{s}$ from sensory 
  input $\boldsymbol{y}$. The \textcolor{lrc_color}{\lrc}
  immediately translates this world state into a local policy $\boldsymbol{\pi^l}$.
  The \textcolor{cmo_color}{continuous motion optimization} computes a plan $\boldsymbol{\pi^g}$ for some time-horizon
  at a slightly lower rate. \textcolor{rp_color}{\Rp} combines these two
  policies into one policy $\boldsymbol{\pi}$, which enables it to immediately react to local changes, and to look 
  ahead in time to react to larger changes. Finally, this policy $\boldsymbol{\pi}$ produces 
  a control output $\boldsymbol{u}$ (omitted for readability) for  which is sent to the robot.
  \label{fig:system}
  }
\end{figure}

\begin{figure*}[!tbh]
  \centering
  \hspace{-4mm}
  \begin{tikzpicture}
    \node (image1) at (0,0) {
      \includegraphics[width=0.245\textwidth,trim={0 10pt 0 55pt},clip]{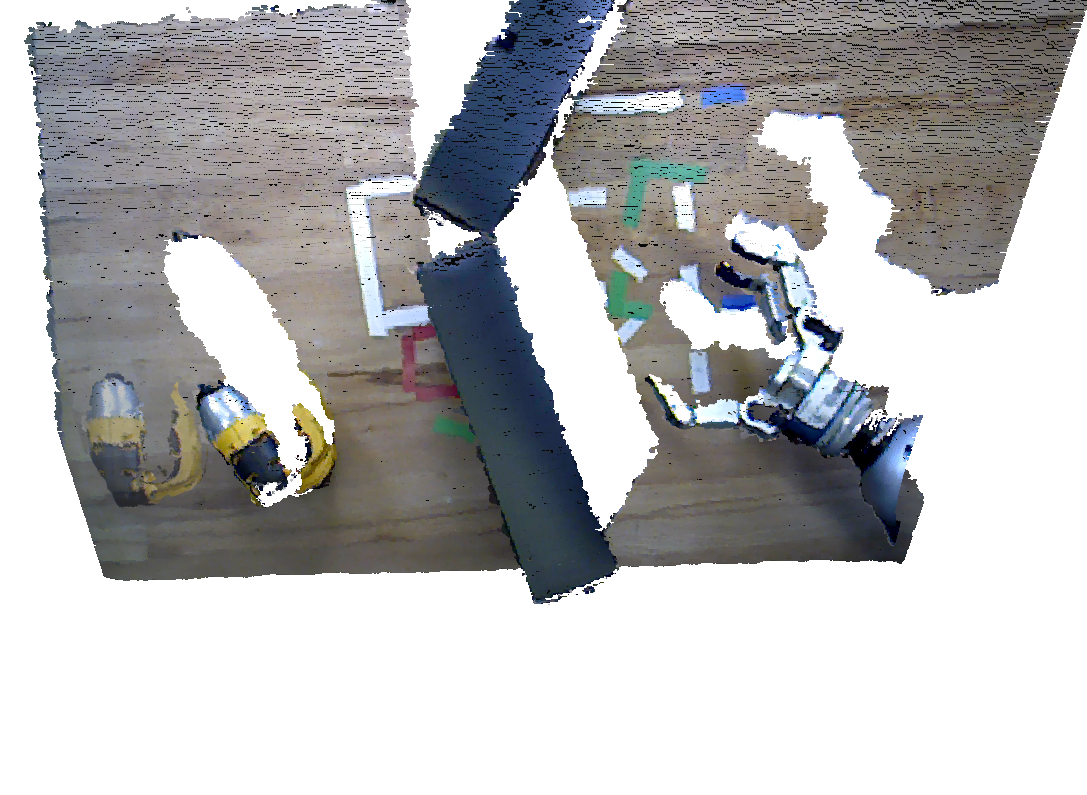}};
    \node[align=left, yshift=-4mm, xshift=8mm] at (image1.north west) {$\boldsymbol{y}$};
  \end{tikzpicture}\hspace{-3mm}
  \begin{tikzpicture}
    \node (image1) at (0,0) {
      \includegraphics[width=0.245\textwidth,trim={0 10pt 0 55pt},clip]{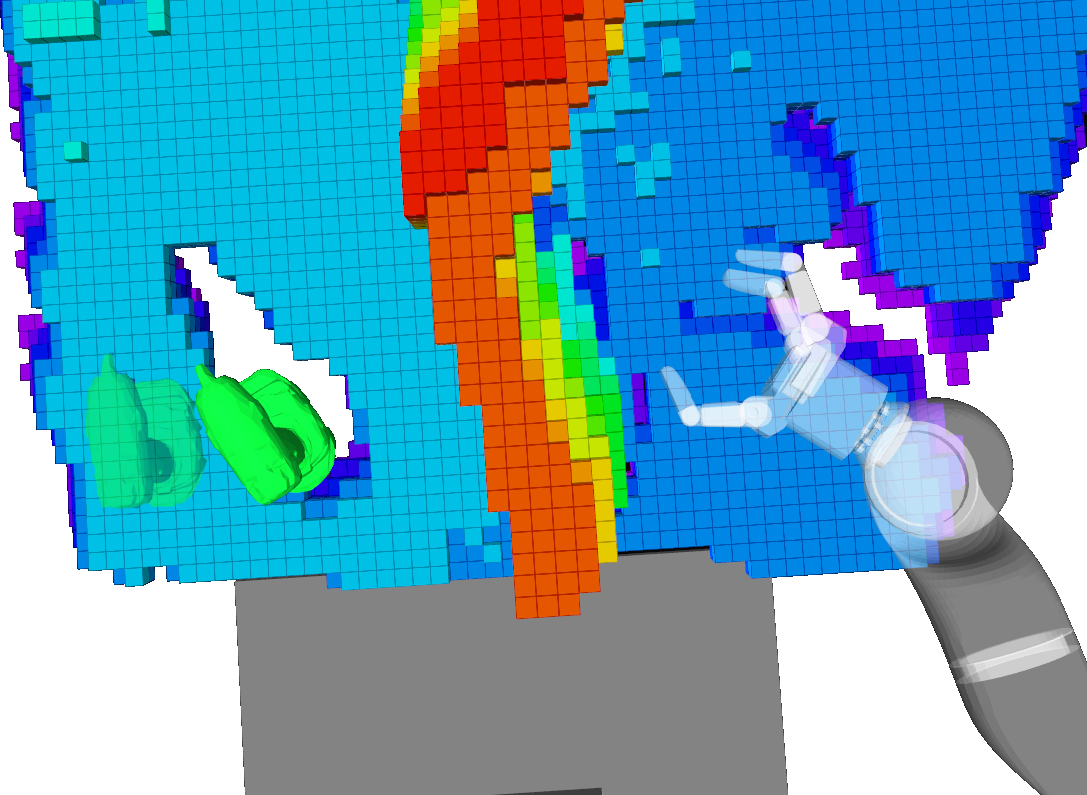}};
    \node[align=left, yshift=-4mm, xshift=8mm] at (image1.north west) {$\boldsymbol{s}$};
  \end{tikzpicture}\hspace{-3mm}
  \begin{tikzpicture}
    \node (image1) at (0,0) {
      \includegraphics[width=0.245\textwidth,trim={0 10pt 0 45pt},clip]{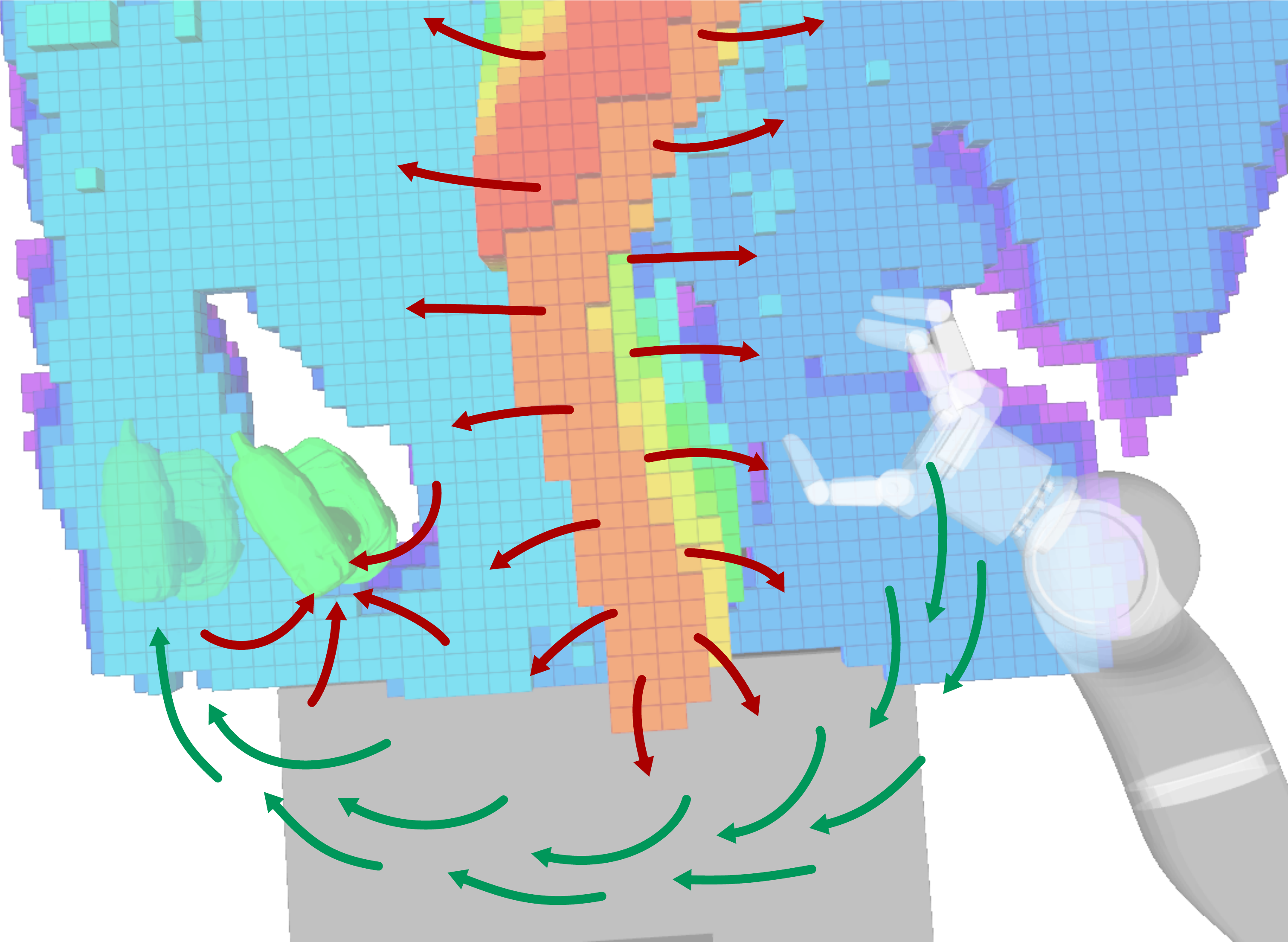}};
    \node[align=left, yshift=-4mm, xshift=8mm] at (image1.north west) {$\boldsymbol{\pi^l},\boldsymbol{\pi^g}$};
  \end{tikzpicture}\hspace{-3mm}
  \begin{tikzpicture}
    \node (image1) at (0,0) {
      \includegraphics[width=0.245\textwidth,trim={0 10pt 0 45pt},clip]{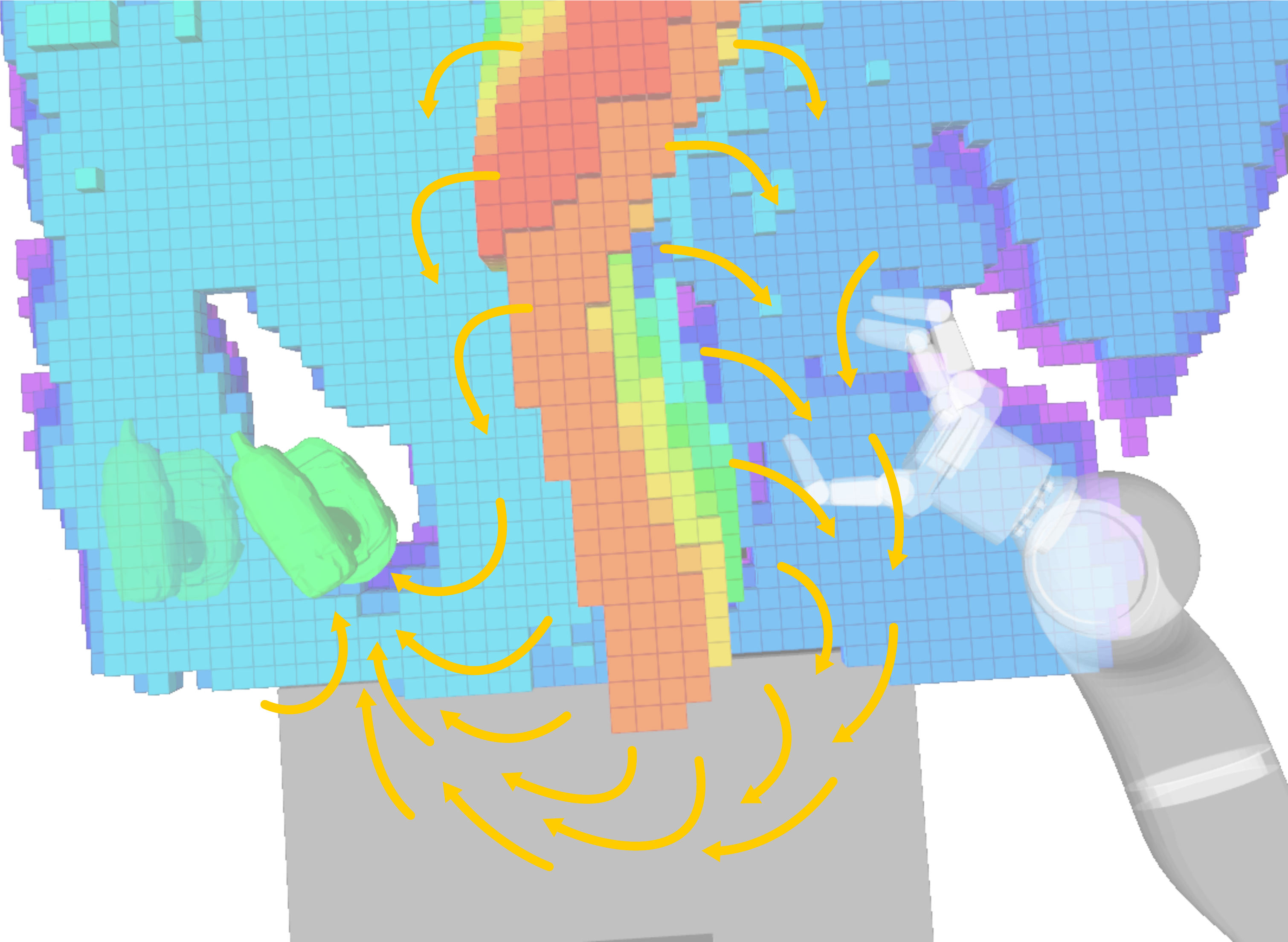}};
    \node[align=left, yshift=-4mm, xshift=8mm] at (image1.north west) {$\boldsymbol{\pi}$};
  \end{tikzpicture}
  \caption{Illustration of one time step of Fig.~\ref{fig:system} with
    the same color coding. From left to right: sensory input
    $\boldsymbol{y}$ (we overlay the position of target object at an
    earlier time step), perceived state $\boldsymbol{s}$ of the robot
    and the environment (target object, obstacles), local and global
    policies $\boldsymbol{\pi^l}$, $\boldsymbol{\pi^g}$, and fused
    policy $\boldsymbol{\pi}$. }
  \label{fig:system-pics}
\end{figure*}

\section{System Architectures}\label{sec:archs}
We evaluate and compare three alternative system
architectures along the spectrum from feedback control to motion
planning~\cite{apc_2016}: (i) \SPA, (ii) \LRC~and (iii) \RP. We are
aware that these names still contain some ambiguity.
In the absence of existing terminology we define what we mean by them
in each subsection and use them coherently in this paper. 
Figs.~\ref{fig:system} and~\ref{fig:system-pics} present an overview
of how the information flows between the perception and the motion
generation modules in the different architectures.
We discuss them here in relation to related work on robotic
systems. A review of the vast body of
work on visual tracking or motion planning is out of the scope of this
paper and we refer to the respective
sections in~\cite{pot,2016_RAL_armtracking,Ratliff:15,2016_IROS_jim}.

\subsection{\SPA}
Building systems through strong modularization into sensing, planning
and acting components remains the predominant paradigm for high DoF
robotic system building ~\cite{KortenkampSB16}. In this paradigm,
perception provides a model of the environment, in which a motion
planner finds an optimal, collision-free path that is then tracked by
a stiff and accurate controller. In Fig.~\ref{fig:system}, this
architecture corresponds to perception (in blue) and motion
optimization components (in green), with visual feedback being
considered only at the beginning of the motion planning task.

The advantage of this paradigm lies in the subdivision of the complex
problem of robotic manipulation into intuitive subproblems that are
easier to solve. Systems that are built according to {\em
  sense-plan-act\/} (SPA) are perfectly suited for environments that
are well-defined, structured and controlled.
However, they cope less well in the presence of uncertainty and a
changing environment~\cite{APC_AllTeams}. Due to the well known
limitations of {\em sense-plan-act\/}~\cite{Brooks:1990}, robotics
researchers have proposed extension, e.g. {\em sequential
  sense-plan-act\/} (seqSPA), acknowledging the importance 
of environmental feedback during motion execution. Here, the robot
does not only request feedback once at the very beginning but also at
deliberately chosen moments during task execution. 
This approach is more robust against uncertainties in both sensing and
actuation than SPA. However, it is still not able to cope with fast
environment or target object dynamics. Some teams competing in the APC
and in the DARPA ARM Challenge followed
seqSPA~\cite{APC_AllTeams,Righetti2014ARM}.

\subsection{\LRC}
On the other end of the spectrum from feedback control to motion
planning, we consider system architectures that
rely entirely on visual feedback control. They do not have the global
motion optimization modules (green in Fig.~\ref{fig:system}), but
purely rely on local policies (red in Fig.~\ref{fig:system}) to
generate reactive motion behavior. With {\em local\/}, we indicate
that they only take the local geometry of the environment around the
current manipulator pose into account to compute the optimal,
immediately next control command. System architectures in this
category react to changes immediately and are very robust to
uncertainties in sensing and actuation. However, they may get stuck in
local minima for example when the environment has a complex workspace
geometry~\cite{Kuffner2016}.

Systems that are entirely based on feedback control have a long
tradition. For example, \cite{khatib1986real} proposes a well defined
interface for perceptual feedback in form of potential fields
constructed from closest points. The resulting feedback control law
can be computed efficiently using the superposition property of
potential fields in combination with additive control laws based on
the desired motion and constraints such as joint limits.

Visual servoing~\cite{ChaumetteSB16} broadly refers to the
class of \lrc\ that closes the loop around visual data.
More recently, there has been a lot of work on learning motion
policies directly from perceptual feedback in form of raw camera
images and the system joint state, e.g.~\cite{Levine:2016}.
Another example comes from the team who won the first
APC~\cite{apc_2016}. Although they do not close the loop around visual
data, they demonstrate the robustness of standard
joint space and operational space controllers. The 
authors admit that \lrc\ approach may have limitations in more complex
manipulation tasks that require planning.

\subsection{\RP}
Summarizing the above, \lrc\ gives robotic systems the ability to
immediately consume new information, instantly react to changes and
compensate for inaccuracies. However, it is susceptible to local
minima. Motion planning as typically used in \spa\
finds solutions even in complex situations where feedback
controllers may get stuck. However, this comes at a significant
computational cost that may break real-time requirements.

Ideal would be a hybrid system that combines both, reactive motion
planning and locally reactive control. Such a system can
simultaneously adapt locally but also re-plan in case of larger
changes. Compared to SPA, these systems are much more reactive and
faster in completing the manipulation tasks. They 
rely on two motion generation modules, as depicted in
Fig.~\ref{fig:system}: the global motion optimization 
(green) and local policies (red). The motion representations of both
modules need to be fused to generate one policy for motion generation
(yellow).

Combining local control with motion planning is quite common in the
area of mobile robots, e.g.~\cite{ROB:ROB20109,5175292,
  leonard2008perception,Richter2016}. However, fewer approaches exist to date that
scale up such a hybrid system to robots with many degrees of freedom
which are manipulating the world.

One example is the elastic-strip framework \cite{brock2002elastic}
which combines local control with motion planning. While it conceives
the use of on-board vision sensing, it is demonstrated on a real
platform but with a simulated, potentially changing world model. 
Controller funneling~\cite{burridge1999sequential} is another hybrid
approach which takes perceptual information into account.
This approach requires a-priori knowledge to design
the state space partitioning controllers and their switching
conditions.

{\em Dynamic Movement Primitives\/} (DMPs)
\cite{IjspeertPastorDMPs2013} can be interpreted as a combination of
local control and planning. Feedback terms can be learned and
incorporated~\cite{rai2016learning} for instantaneous reaction.
Furthermore, perceptual feedback can be used to dynamically switch
DMPs \cite{Kappler-RSS-15}, which in turn results in local reactive
control policies.

\cite{Lehner2015} presents a mobile manipulation system that locally
adapts and augments global motion plans in response to changes in the
environment as perceived by on-board sensors.
\cite{7803288} present a system to find valid stance and
collision-free reaching configurations in complex, dynamic
environments for a full humanoid robot. 

In this paper, we compare instantiations of each of these
three different architectures that consist of the same components. In
the next sections, we briefly describe these components and the
interface between them. Sec.~\ref{sec:exp} presents
experimental results that are then discussed in
Sec.~\ref{sec:conclusion}. 

\section{Real-Time Feedback Modules}\label{sec:feedback}
We consider scenarios that require manipulation of
dynamic objects in dynamic environments. Continuous feedback on the
location of target objects and the workspace structure is of utmost
importance for systems acting in these scenarios. An important
requirement for the feedback components is therefore to deliver
information as fast as possible. A low-dimensional representation of
this information is beneficial to keep bandwidth requirements low.
 
In the following, we briefly describe the methods integrated into our
system. Any other method which fulfills the requirements could be used
instead.  

\subsection{Visual Tracking of Target Objects}
We use visual tracking to estimate the pose of every object the robot
may want to manipulate.
We choose the probabilistic method from our previous work \cite{pot}. 
This method assumes knowledge of the 3-dimensional shape of the
objects of interest, represented as triangle meshes.
It takes as input depth images and compresses it into
\textit{6 DoF object poses} at the frame rate of the on-board camera.
Its formulation makes it very robust to occlusions of object parts
which are common in the context of manipulation tasks. 

\subsection{Visual Robot Tracking}
Precise positioning of a robot arm with respect to the sensed
environment and target object is a crucial
ability for manipulation systems. This is not always
possible through naive application of forward kinematics. On real
robotic platforms, kinematic models and measured joint angles are
commonly inaccurate and therefore lead to erroneous predictions of
end-effector pose relative to the camera.

To mitigate this problem, we continuously estimate the true robot arm
configuration relative to the camera mounted on the robot's head. We
choose the probabilistic, real-time method from our previous work
\cite{2016_RAL_armtracking}.
It fuses depth images and measured joint angles to produce precise
estimates of the \textit{robot configuration} at 1kHz which is the rate
of the joint encoders.
Even under heavy occlusion of the arm and very fast motion, this
method can correct errors due to biases in the joint sensors and to
imprecise kinematics of the camera relative to the rest of the
kinematic chain. Furthermore, it models the delay between the
measurements from the joint sensors and the camera.

\subsection{Modeling Unstructured Workspace Obstacles}
To generate collision-free motion, the robot needs to be aware of the
workspace geometry and the obstacles therein.
We reconstruct this geometry from depth images and represent obstacle
regions in a discrete occupancy grid.

Commonly, occupancy of the environment is represented
probabilistically and at multiple scales, e.g. using
\cite{hornung2013octomap}. This however, comes at a significant
computational cost. Instead, we process depth images frame-by-frame,
which allows to generate an occupancy grid at approximately 15Hz.
Empirically, we found this rate to be sufficient. For each frame a
single-scale, noise-filtered voxel grid is generated. The grid is
cropped to the reachable robot workspace. Points corresponding to the
tracked robot arms and the tracked object are removed, see
Fig.~\ref{fig:system-pics}. The occluded regions are set occupied
using ray casting \cite{hornung2013octomap}.

\section{From Visual Feedback to Continuous Signed Distance Fields}
Given the processed perceptual data as described in the previous
sections, we convert it into a set of {\em Signed Distance
  Fields\/} (SDFs) describing the target object, table, unstructured
workspace obstacles, and the robot.

\begin{figure}[t]
      \centering
      \includegraphics[width=0.6\linewidth,trim={0 300pt 0 0},clip]{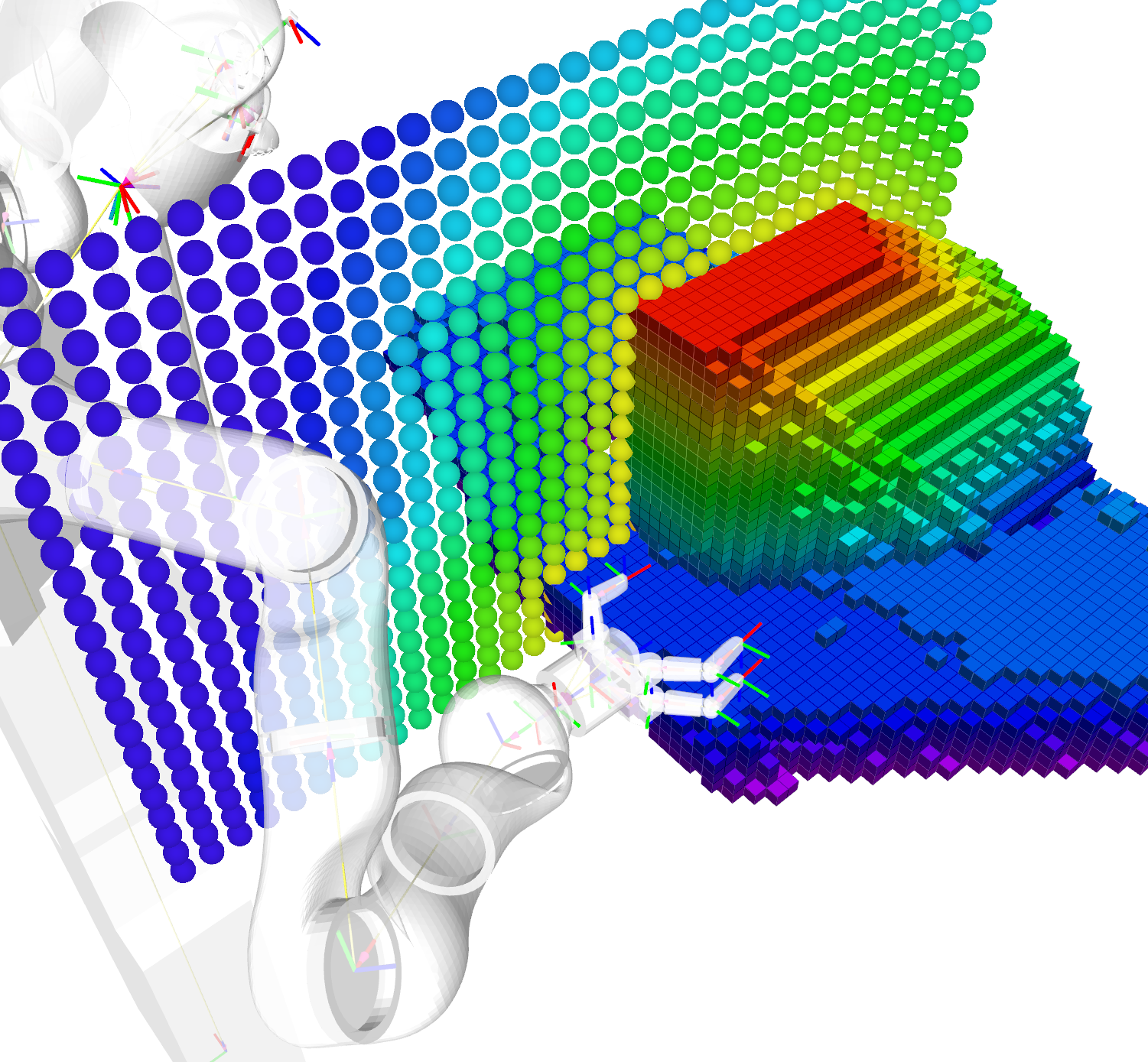}
      \caption{The occupancy grid (cubes: colors represent height), 
        along with a vertical slice of the corresponding signed distance field 
        (spheres: color represent the distance to the closest occupied cell).}
      \label{fig:sdf}
\end{figure}

SDFs also provide a way to perform fast collision
checking in motion planning and have previously been used as the
underlying data structure for motion optimization algorithms
\cite{Zucker:13, Ratliff:15, Kalakrishnan:11}.
SDFs also allow to define proper Riemannian metrics to measure path
length in workspaces populated by obstacles \cite{Ratliff:15}, which we
make use of in \rp. We use local interpolation
techniques~\cite{2016_IROS_jim} to define gradients of the discrete
SDF from the occupancy grid map (see Fig.~\ref{fig:sdf}).

We approximate each target object, table and robot body by simple
geometrical shapes like spheres, boxes and capsules (see
Fig.~\ref{fig:avoidance_controllers}) to allow efficient distance
computation. We position the SDF of the target object according to its
visually estimated 6DoF pose. The robot arm configuration is updated
according to the joint angles as estimated by the aforementioned
sensor fusion approach. We assume a known, static pose of the table.

Sending the full SDFs to \lrc\ would require high bandwidth
communication. At the same time the full workspace information cannot
be leveraged in the fast 1 KHz loop.
Hence, we further approximate the workspace geometry with \textit{subsets of
closest points} for each component of the approximated robot model (see
Fig.~\ref{fig:avoidance_controllers} (right)).
Through their derivative, each distance field locally encodes the
location of the closest point in the component of the environment that
it is concerned with.
We use this property in \lrc.

\section{Motion Generation}\label{sec:motion}
The motion generation module is designed for continuous feedback
integration. The architecture enables blending local reactive control
and higher-level continuous motion optimization. More specifically, we
use reactive control policies of the form $\pi_A = (f,{\bf A})$ that
are defined as a non-linear second-order differential equation of the
form $\ddot{\bf x} = f({\bf x}, \dot{\bf x})$ with
${\bf x}, \dot{\bf x}, \ddot{\bf x} \in \mathbb{R}^k$ and a positive
semi-definite weighting matrix that may vary smoothly across the space
\mbox{${\bf A}:\mathbb{R}^k \times \mathbb{R}^k \rightarrow
\mathbb{R}^{k\times k}$}. Suppose we have a number of task spaces
${\bf x}_1, \ldots, {\mathbf x}_n$ defined by differentiable maps
${\bf x}_i = \phi_i({\bf q})$ from configuration space such as forward
kinematics maps to points on the robot's body, maps to relative
distances between the robot and workspace objects
\cite{toussaint2010bayesian}. Then given a collection of local
controllers $(f_i, \phi_i, {\bf A}_i)$ defined on those task spaces,
we define their pointwise evaluation at joint angles ${\bf q}$ as
\begin{equation}
\ddot{\bf q}^d =\argminA_{\ddot{\bf q}} \frac{1}{2} \sum^n_{i=1}
||{\ddot{\bf x}}_i^d - {\bf J}_i{\ddot{\bf q}} ||^2_{{\bf A}_i}
\label{eq:quad_prog}
\end{equation}
where ${\ddot{\bf x}}_i^d = f_i({\bf q}, {\dot{\mathbf q}})$ and
${\bf J}_i$ is the Jacobian of $\phi_i$. Here, we use a common
Gauss-Newton-like approximation, removing the second-order term in the
expression
$\ddot{\bf x} = {\bf J} {\ddot{\bf q}} + {\dot{\bf J}} {\dot{\bf q}}$.

The local collision avoidance
controllers can be derived from complex motion policies that are
optimized over a longer time horizon, e.g. as LQRs (see below in
Sec.~\ref{sec:motion_gen_reactive_planning}). This modularity
enables us to experiment with the full spectrum from locally reactive
control to higher level motion optimization for longer horizon
reasoning.

\subsection{\LRC}
\label{sec:motion_gen_feedback_control}
\begin{figure}[t]
  \centering
   \includegraphics[height=70pt]{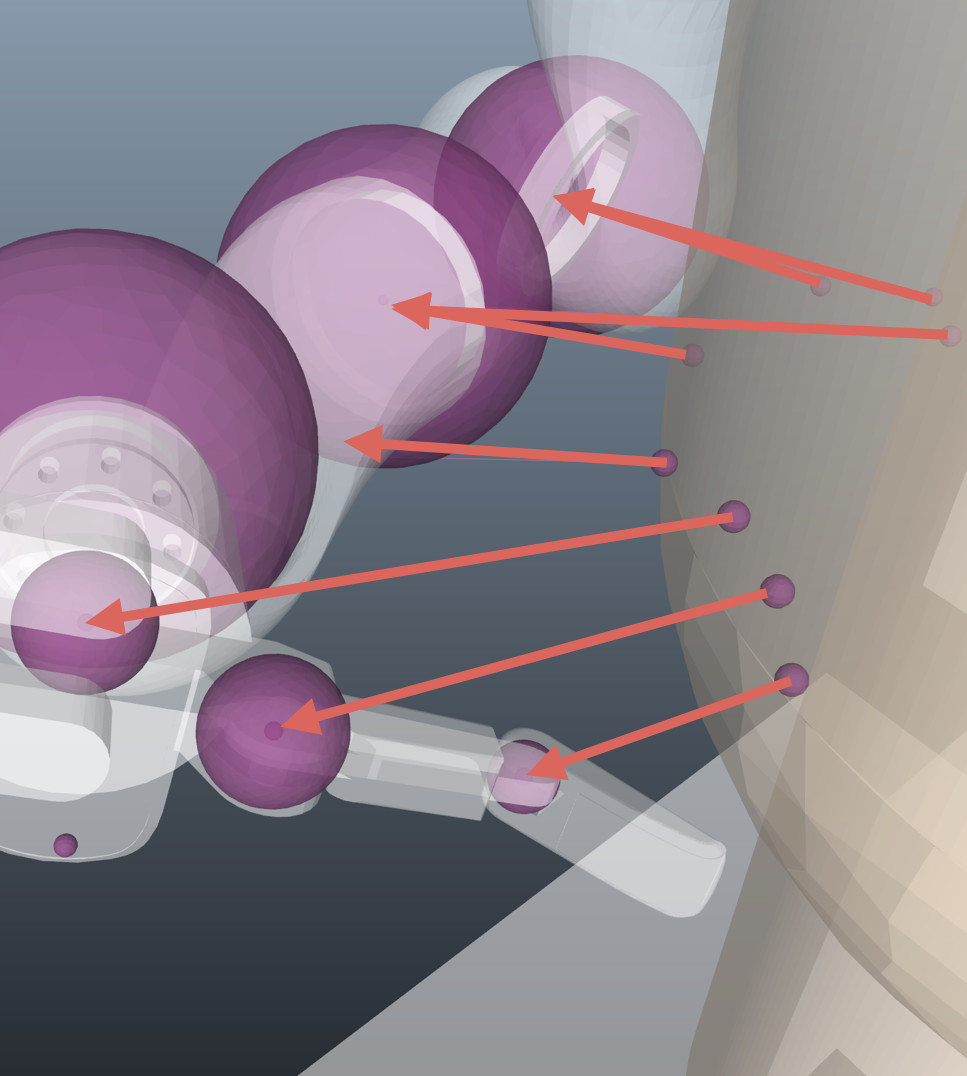}
   \includegraphics[height=70pt]{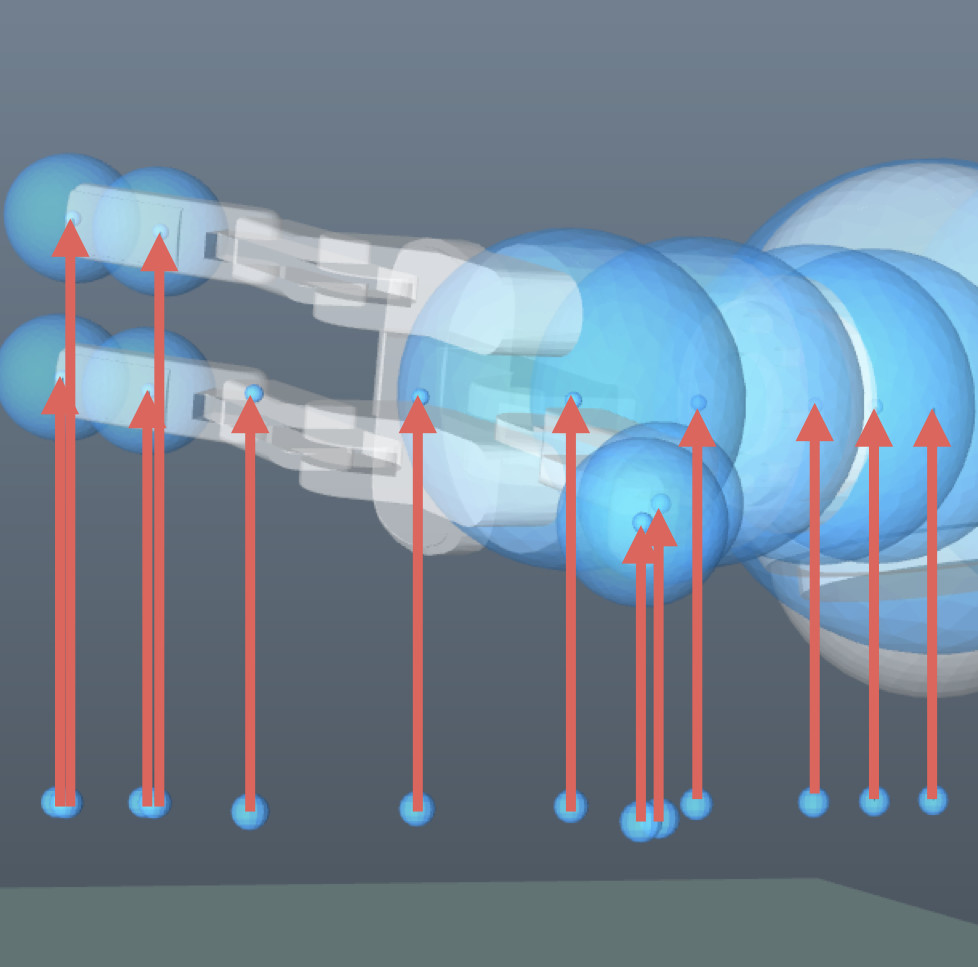}
   \includegraphics[height=70pt]{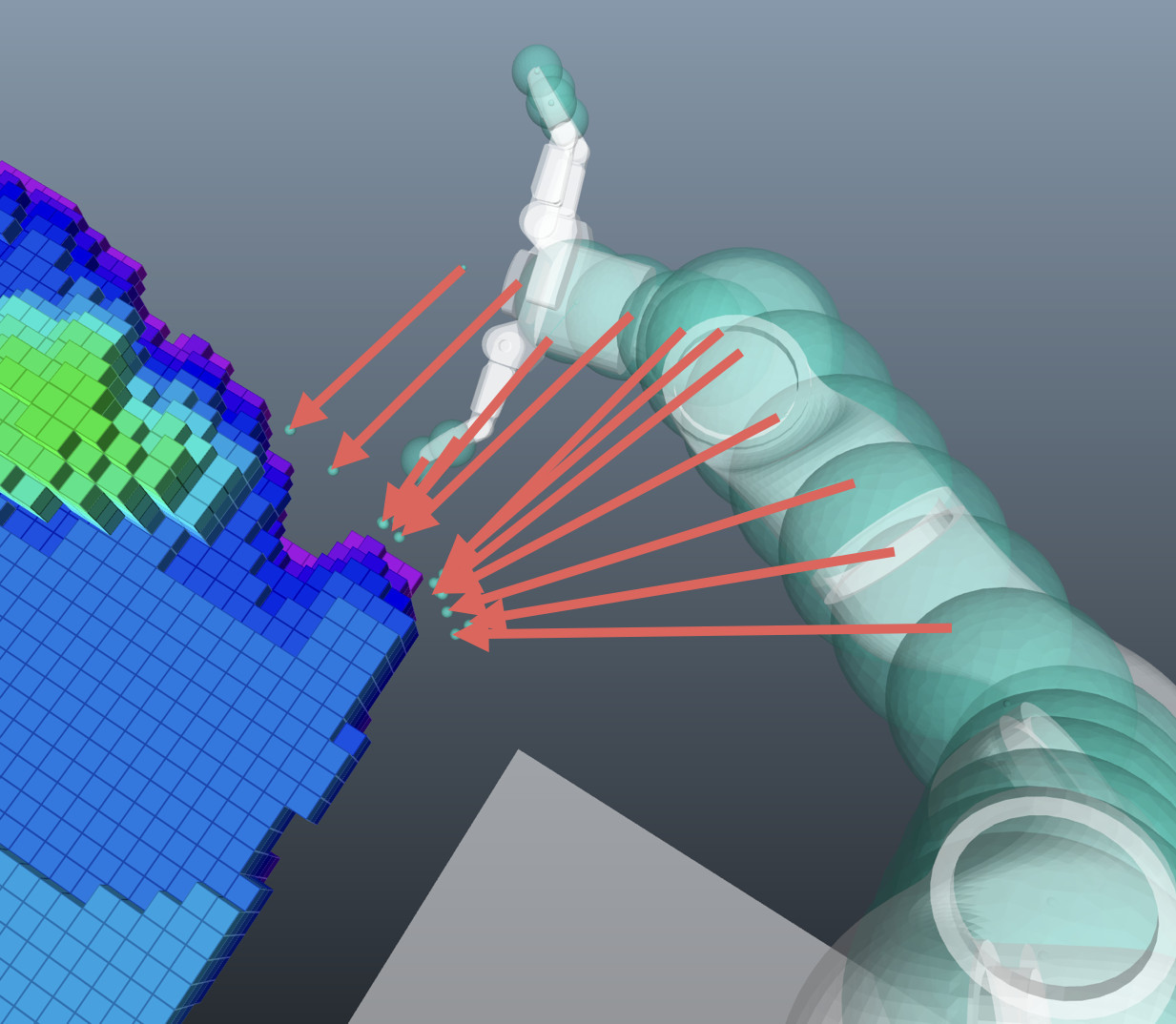}
   \caption{\label{fig:avoidance_controllers}
     Local avoidance controllers to robot (left), table
     (middle) and ambient world (right).}
\end{figure}

\LRC\ combines multiple controllers through Eq.~\ref{eq:quad_prog},
including collision controllers to instantaneously react to the local
workspace geometry, and target controllers for goal convergence. The
target controller pulls the system toward position and orientation
targets in a purely local Cartesian control fashion. This portion of
the system can therefore be used by itself, without any higher level
planning. It is visualized in Fig.~\ref{fig:system-pics} with red
arrows. In addition to the target controller, we use a collection of
obstacle avoidance controllers that take effect when parts of the body
get close to an obstacle. They create workspace accelerations away
from the obstacle with increasing priority as a function of proximity.
They are visualized in Fig.~\ref{fig:avoidance_controllers}.

We also use a default posture potential that pulls the arms slightly
toward a default posture to resolve redundancy, along with simple
damping controllers in the c-space and at the end-effector to regulate
the velocity of the system.

Local control runs at 1kHz for effective integration of the underlying
highly nonlinear differential equation. However, it sends joint
positions, velocities and accelerations for low-level execution only
at 100Hz. The lowest level of control handles the generation of the
torques needed to track the desired joint states through interpolation
and an inverse dynamics controller. 

\subsection{\RP\ and \CMO}\label{sec:motion_gen_reactive_planning}
On the other end of the spectrum, we use a motion optimizer based
on Riemannian Motion Optimization (RieMO) \cite{Ratliff:15}. It runs
continuously, tracking the local minimum  
based on feedback while obstacles and the target change over time.
This motion optimizer integrates information over a time horizon of
three seconds (the (approximate) average time length
of a reaching motion), enabling anticipatory behaviors and efficient
coordination of collision controllers and potentially multiple target
controllers. As is done in optimal control and MPC, it summarizes its
policies as Linear Quadratic Regulators (LQRs) built on a local
quadratic approximation around the local optimum, however only
kinematically (with accelerations as actions) since the planning
module addresses only movement. This is visualized in
Fig.~\ref{fig:system-pics} with green arrows. These are sent to
\lrc~for integration with the other controllers through
Eq.~\ref{eq:quad_prog}. The continuous optimizer operates at a slower
time-scale than \lrc, updating its optimization at 5-10Hz. To mitigate
potential delays, it sends full LQR policies that represent the
optimal policy within a region of the locally optimal trajectory.
Rather than using a simple attractive potential pulling the
end-effector toward a desired pose (which can be expressed as a
differential equation in the configuration space) the motion optimizer
creates a more expressive attractor differential equation that
simultaneously pulls the system toward the desired target while also
integrating anticipatory actions that enable smoother, more efficient,
and well-coordinated behavior. Therefore, the local controllers are
able to operate cohesively with the planned policies between planning
updates.

\subsection{Grasping}
We decompose the grasp problem into multiple sequential task
states---approach, establish grasp, move object, release, and
retract---each governed by either local control, continuous
optimization, or some combination thereof. Grasping is controlled by
independent controllers, while the rest of the system observes the
resulting movement of the hand and reacts accordingly to
simultaneously adjust the arm to avoid obstacles and stabilize the
hand posture to the extent possible under the constraints of the
environment. This enables us to setup consistent experimental
scenarios for empirical study. We manually defined a set of grasp
poses for each object that we use in our experiments.

\section{Experiments and Demonstrations}\label{sec:exp}
We compare three different system architectures that were explained in
more detail in Sec.~\ref{sec:archs}: (i) \spa, (ii) \lrc\ and (iii)
\rp.
As an experimental platform, we use a fixed-base, manipulation
platform equipped with two $7$-DoF Kuka LWR IV arms, three-fingered
Barrett Hands and an RGB-D camera (Asus Xtion) mounted on an active
humanoid head by Sarcos. All components are torque controlled using an
inverse dynamics controllers to track the desired joint states. It
runs at determinable worst-case execution times of 1ms and is executed
on a PC running Xenomai, a real-time framework for Linux. 
\subsection{System Architecture Realizations}
The visual perception modules (Sec.~\ref{sec:feedback}) consist of
a tracker for the right robot arm, and a tracker for the target
object. We assume a known table pose. Everything else in the
environment is considered to be unstructured workspace obstacles
modeled by an occupancy grid map. The algorithms used in all our
implementations of the different architectures are identical. We
vary the frequency at which information is passed to motion generation
and whether we consider policies that are optimized over a longer time
horizon. 

\paragraph{\SPA}
Here, we acquire just one depth image in the very beginning of the
experiment. Based on this image, the poses of the objects of interest
are estimated, and a model of the workspace geometry is created. Then
a one-shot motion optimizer, a simple variant of
Sec.~\ref{sec:motion_gen_reactive_planning}, generates a plan which
will be executed without any further visual feedback. The overall
planning time of \spa\ is limited to 2s (chosen empirically) for all
experiments.

\paragraph{\LRC}
In this architecture, depth images are processed continuously to
estimate the object pose and robot arm configuration. Additionally,
the world model is updated online. This information is consumed by
\lrc\ (see Sec.~\ref{sec:motion_gen_feedback_control})
that immediately adapts to the observed changes in the next control
cycle.

\paragraph{\RP}
As in the previous architecture, the object and robot arm tracker
continuously estimate the object pose and robot arm configuration.
Also the world model is updated online. However, here the information
from the perception modules is also used to continuously replan in
addition to the \lrc\ (see Sec.~\ref{sec:motion_gen_reactive_planning}).

\subsection{Scenarios}
\begin{figure*}
\centering
\begin{subfigure}{.22\textwidth}
  \centering
  \includegraphics[width=.80\linewidth,trim={0 0 0 130pt},clip]{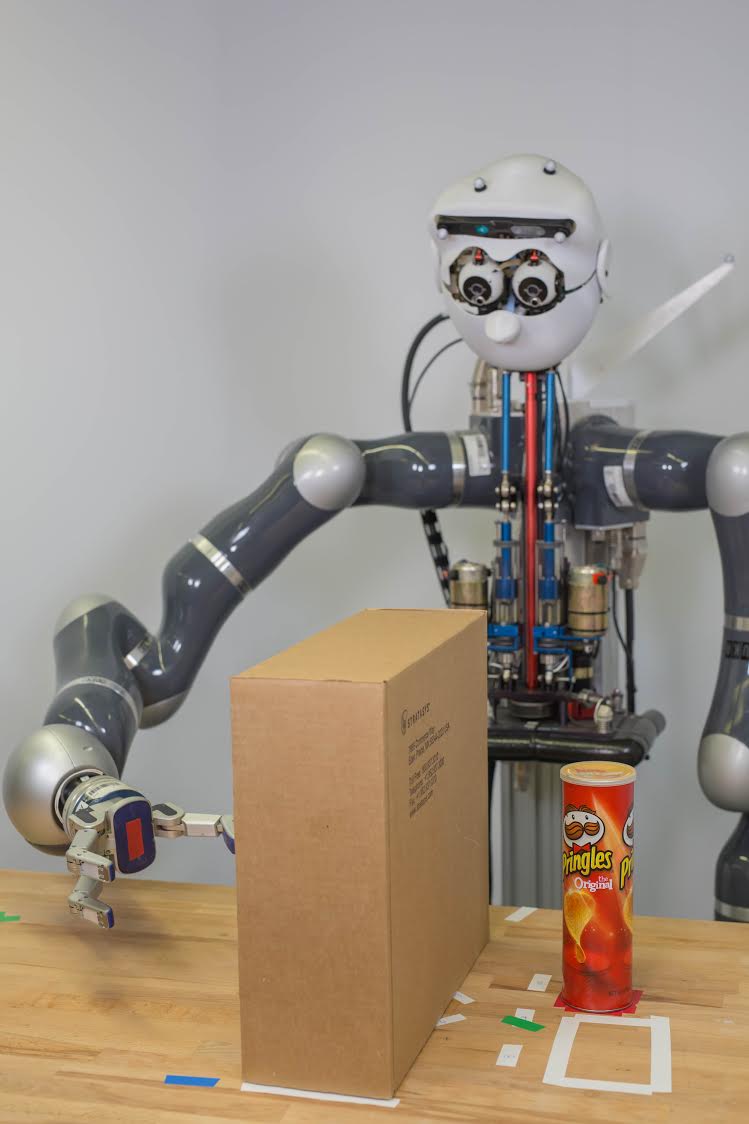}
  \caption{Static pick and place}
  \label{fig:setup_pick_and_place_static}
\end{subfigure}%
\begin{subfigure}{.22\textwidth}
  \centering
  \includegraphics[width=.80\linewidth,trim={0 0 0 130pt},clip]{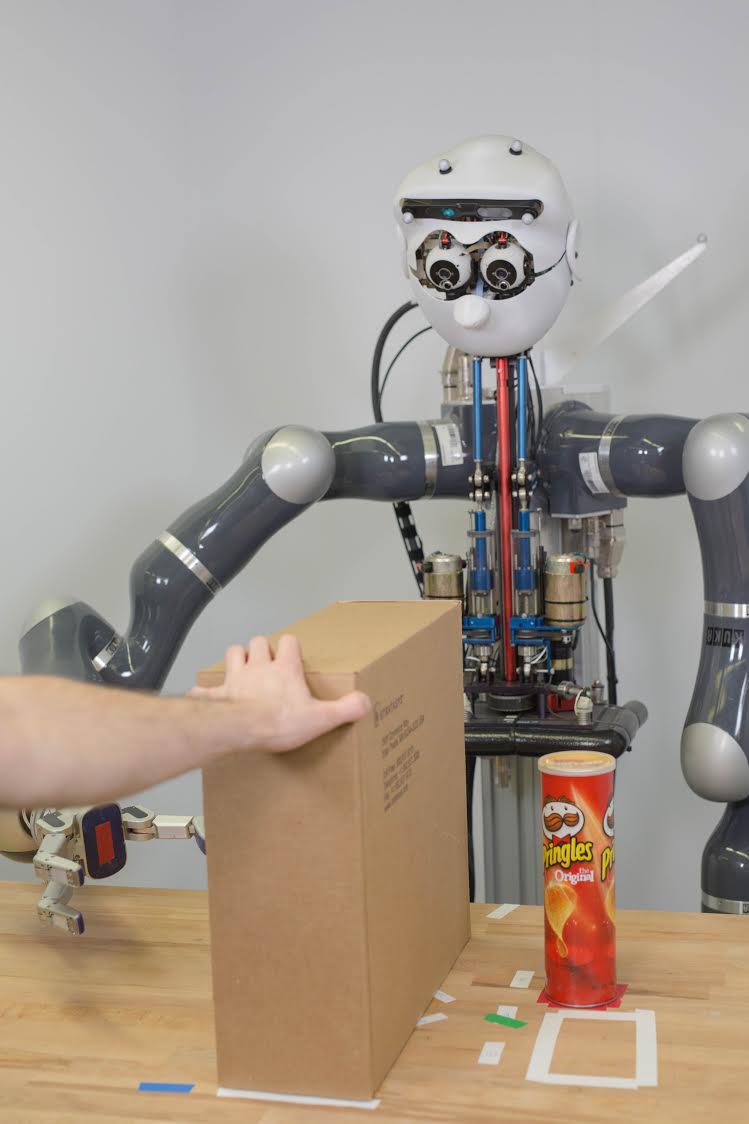}
  \caption{Dynamic pick and place}
  \label{fig:setup_pick_and_place_dynamic}
\end{subfigure}%
\begin{subfigure}{.22\textwidth}
  \centering
  \includegraphics[width=.80\linewidth,trim={0 0 0 130pt},clip]{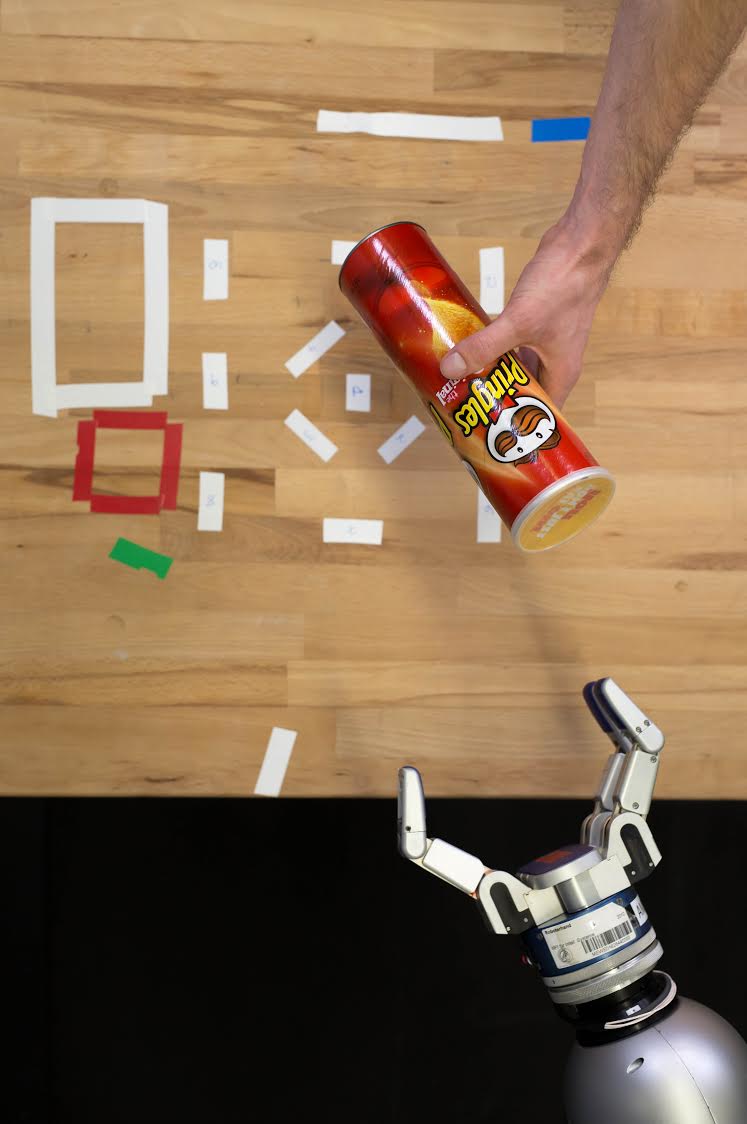}
  \caption{Dynamic grasping}
  \label{fig:setup_grasping_dynamic}
\end{subfigure}%
\begin{subfigure}{.22\textwidth}
  \centering
  \includegraphics[width=.80\linewidth,trim={0 0 0 130pt},clip]{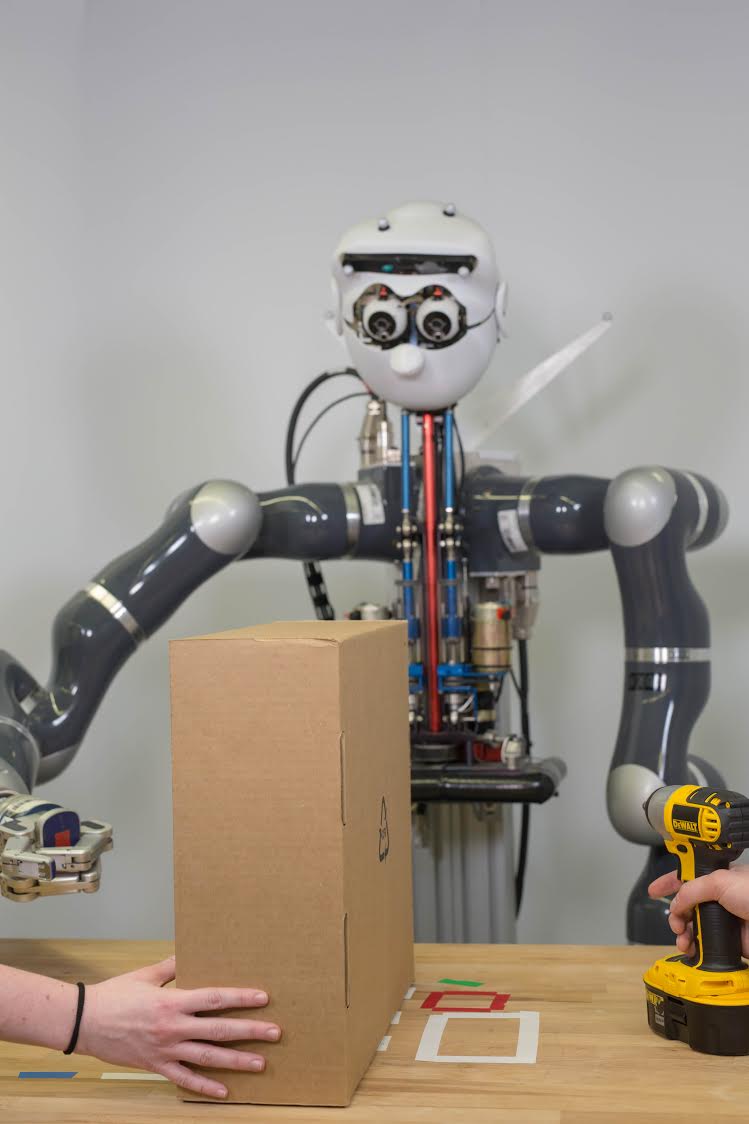}
  \caption{Dynamic pointing}
  \label{fig:setup_pointing_dynamic}
\end{subfigure}%
\caption{Experimental scenarios: The human hands indicate which objects
are being moved during execution.}
\label{fig:test}
\end{figure*}

We present four different scenarios (see Fig.~\ref{fig:test}) and
experimental results which illustrate the importance of tightly
integrating real-time perception and reactive motion generation. Each
experiment instance is performed at least 3 times.

\subsubsection{Pick and Place in Static Environments of Increasing
  Difficulty}
In this experiment we consider the static pick and place scenario
shown in Fig.~\ref{fig:setup_pick_and_place_static}. The task is to
pick up the pringles box and place it on the other side of the brown
box, without any collisions.

\begin{figure}
  \centering
  \includegraphics[width=0.9\columnwidth]{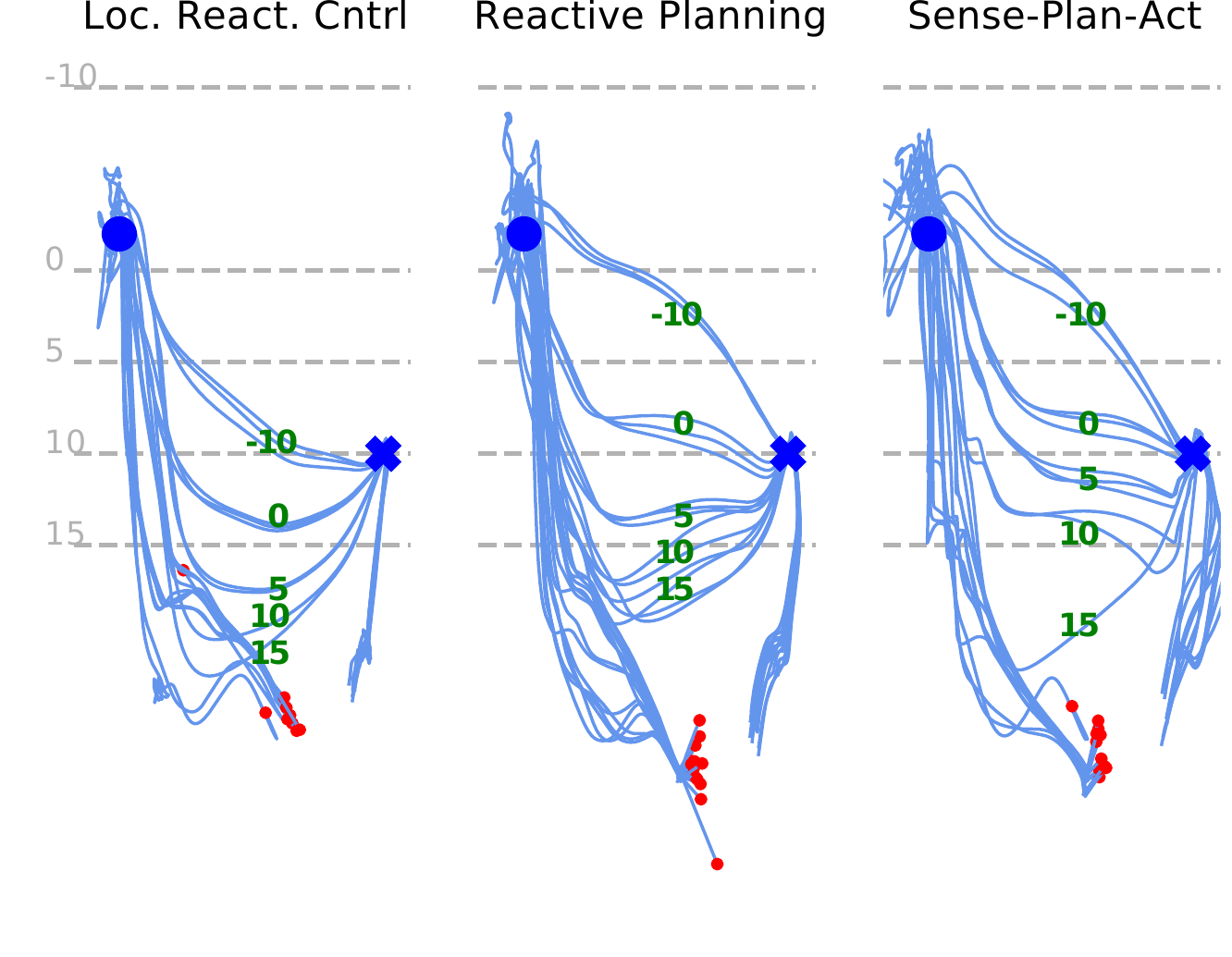}
  \caption{Visualization of the successful end-effector trajectories
    in a pick and place task in the presence of a box obstacle
    (cf. Table~\ref{tab:success_pick_and_place}). Per 
    experiment, the box varies its pose from -10, 0, 5, 10 or 15cm
    distance to the far edge of the table (as indicated by the
    dashed lines). The green labels indicate which trajectory belongs to
    which box position. The red dots indicate the start of each
    trajectory. The blue dot indicated the picking object position and
    the cross its placing position. We compare
    \lrc\ (Left), \rp\ (Middle) and
    \spa\ (Right). 
    \label{fig:static_pick_and_place}} 
\end{figure}

Fig.~\ref{fig:static_pick_and_place} uses gray lines to visualize
the varying positions of the box obstacle in this scenario. The box is
always placed prior to starting each experiment. The closer the box to
the robot base, the higher the difficulty to successfully pick and
place the pringles. For each system architecture and complexity level
we run three trials. At position 15, we reached the point where each
system failed at least once.

%
%
Table~\ref{tab:success_pick_and_place} shows the success rate of
picking up the object and placing it at the target location. Even
though the planning problem itself becomes very challenging,
\lrc\ alone already performs very well. Not surprisingly, 
\spa\ performs very well in such a static environment. However, in the
most challenging setting it does fail more often compared to \rp. One
reason for this is the limited planning time allocated, during which
no successful plan may be found. \Rp\ is able to find a path more
often since it is able to re-plan continuously, thus, has more time to
find a feasible path during execution.

In Table~\ref{tab:time_pick_and_place} we report the average execution
time for successful trials in seconds. Time required for the
initial object detection is not part of the execution. \Lrc\ and \rp\
are on par for the simple settings, whereas \spa\ is significantly
slower. The difference in execution time is because \spa\ can only
start planning to the next pose after it achieved the 
old pose. It then has to wait until a solution has been found.
The execution time increases with difficulty due to more confined
workspace which results in slower convergence especially for 
\lrc\ and in general for longer trajectories. 

%
%
This scenario illustrates that even in static environments, the two
tightly integrated system architectures can have benefits over \spa.

\begin{table}
\centering
\caption{Success rates (total runs) of pick-and-place
  experiment\label{tab:success_pick_and_place}. Difficulty refers
  to obstacle position (cm) relative to the robot (see
  Fig.~\ref{fig:static_pick_and_place}).}
\begin{tabular}{l|r|r|r}%
  \bfseries Difficulty & \bfseries l. react. c. & \bfseries react. pl. & \bfseries s-p-a\\
  \hline 
  \hline 
  static -10 & 100\% (3)& 100\% (3)& 100\% (3)\\
  static 0 & 100\% (3)& 100\% (3)& 100\% (3)\\
  static 5 & 67\% (3)& 100\% (3)& 100\% (3) \\
  static 10 & 33\% (3) & 100\% (3) & 67\% (3) \\
  static 15 & 17\% (6) & 50\% (6) & 17\% (6)\\
  \hline 
  dynamic & 100\% (3) & 100\% (3) & 0\% (3)  
\end{tabular}
\end{table}
\begin{table}
\centering
\caption{Average execution time (seconds) of successful
  pick-and-place\label{tab:time_pick_and_place}. Difficulty
  refers to obstacle position (cm) relative to the robot (see
  Fig.~\ref{fig:static_pick_and_place}).}
\begin{tabular}{l|c|c|c}%
  \bfseries Difficulty & \bfseries l. react. c. & \bfseries react. pl. & \bfseries s-p-a\\
  \hline 
  \hline 
  static -10 & 13.64 s  & 13.41 s & 21.05 s \\
  static 0 &14.29 s & 13.44 s  & 21.76 s \\
  static 5 & 15.83 s & 14.26 s  & 20.16 s  \\
  static 10 & 20.95 s & 18.32 s & 21.51 s  \\
  static 15 & 28.75 s  & 21.06 s & 18.27 s  \\
  \hline 
  dynamic & 17.95 s & 15.44 s & -
\end{tabular}
\end{table}

\subsubsection{Pick and Place in Dynamic Environments}
In this scenario, the target object is placed in the same location as
in the previous scenario, but initially there is no obstacle.
During execution we move an unmodeled obstacle between the
pickup and goal position (see
Fig.~\ref{fig:setup_pick_and_place_dynamic}). The obstacle is
captured by the octomap and thus is considered by both \lrc\ and \rp.

\begin{figure*}
  \centering
  \begin{subfigure}[b]{.33\textwidth}
    \centering
    \includegraphics[width=.9\linewidth]{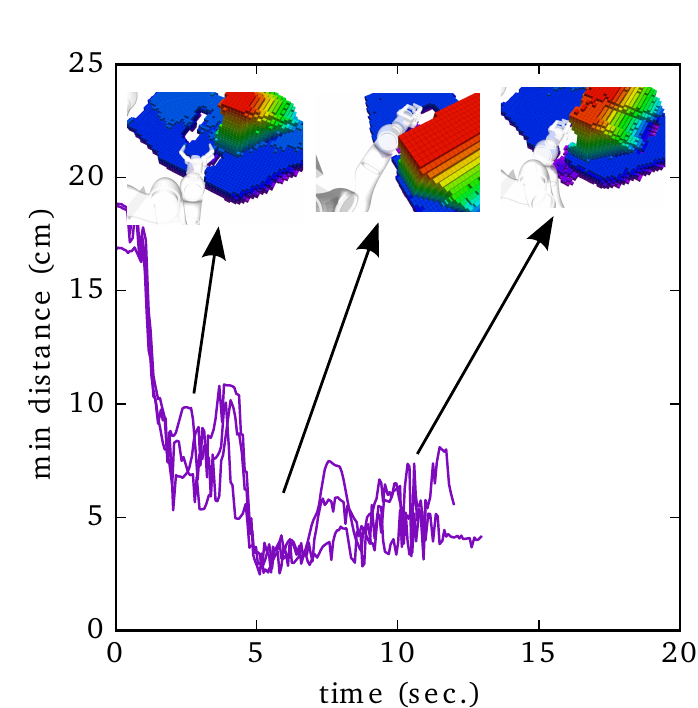}
    \caption{\LRC}
    \label{fig:dynamic_pick_and_place_loc}
  \end{subfigure}
  \begin{subfigure}[b]{.32\textwidth}
    \centering
    \includegraphics[width=.9\linewidth]{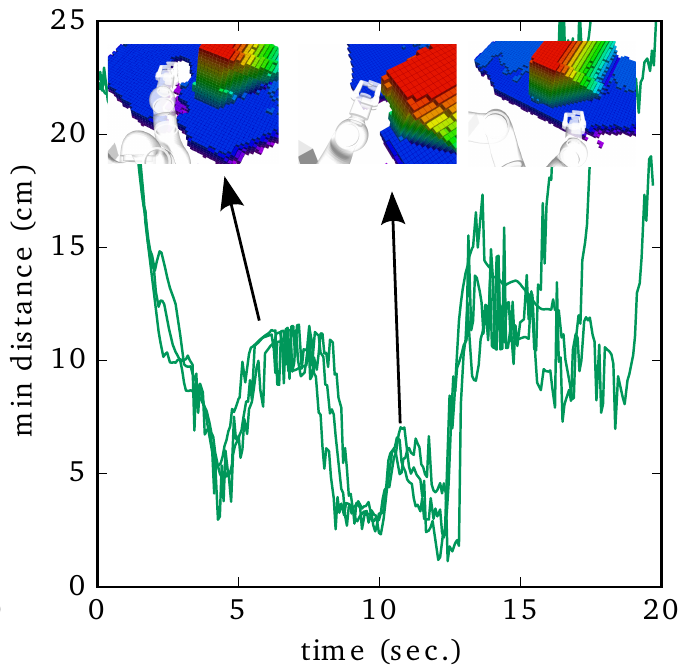}
    \caption{\RP}
    \label{fig:dynamic_pick_and_place_con}
  \end{subfigure}
  \begin{subfigure}[b]{.32\textwidth}
    \centering
    \includegraphics[width=.9\linewidth]{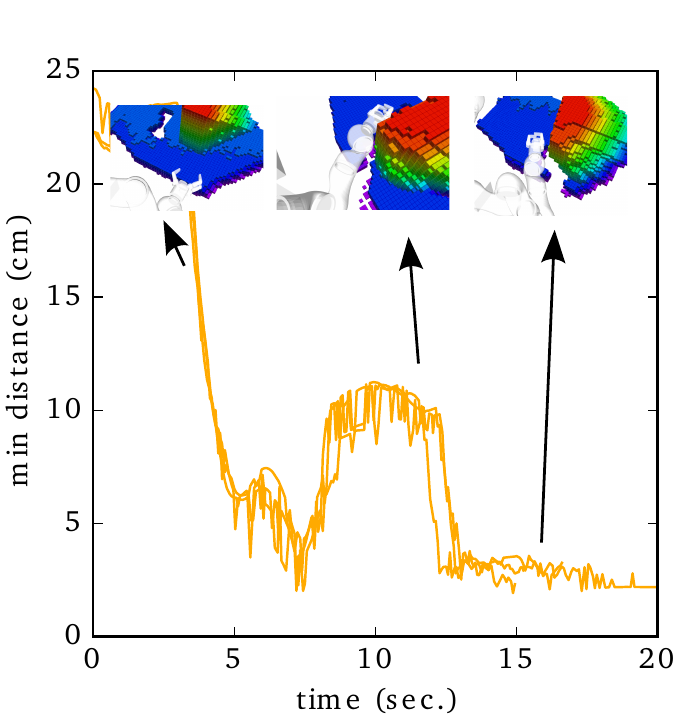}
    \caption{\SPA}
    \label{fig:dynamic_pick_and_place_one}
  \end{subfigure}
  \caption{Minimum distances (i.e, clearance) with the introduced
    obstacle during the dynamic pick and place experiment.
    \Lrc\ gets stuck after successfully grasping the object. \Rp\ while
    slower than the \lrc\ to realize the grasping motion is able to
    perform the place successfully. \Spa\, despite higher initial
    clearance than \lrc\, collides with the obstacle when its position
    changes.}
  \label{fig:dyanmic_pick_and_place}
\end{figure*}

%
%
For the three different system architectures, we observe very
different outcomes. Success rates and execution times are reported in
the bottom row (dynamic) in Table~\ref{tab:success_pick_and_place} and
\ref{tab:time_pick_and_place}. While \lrc\ and \rp\ are on par in
terms of these measures, \spa\ fails completely - as expected. Results
in terms of clearance between the robot and the dynamic obstacle are
summarized in Fig.~\ref{fig:dyanmic_pick_and_place} for 
the nine experimental runs (three per architecture). In the case of \spa\, the
motion to perform the pick keeps high clearance with the obstacle.
However, since \spa\ does not react to changes in the environment
during motion, the arm collides with the unmodeled obstacle when its
position changes. \Lrc\ is able to move faster towards the pick
configuration but this results in lower clearance to the
obstacle. \Lrc\ is then unable to perform the place motion and 
gets stuck in a local minimum. When removing the obstacle it nicely
converges to the goal position. Hence, in this case \lrc\ results in
much safer behaviors than \spa. Finally, \rp\ is able to avoid the
introduced obstacle. Additionally, it successfully finds a path to
circumvent the obstacle and to place the object at the goal position.

%
%
In conclusion, this experiments shows the importance of real-time
perception to avoid collisions with unmodeled obstacles. In
addition, this experiment illustrates that having \rp\ is important to
find collision free paths in complex dynamically changing environments
whereas \lrc\ is safe but gets stuck more easily.

\subsubsection{Grasping with Dynamic Targets}
Not only the unmodeled environment is subject to constant change, but
very often also target objects may move when reaching for them. In
this scenario, we systematically analyze the importance of 
perceptual feedback integration into motion generation, when the
target is repositioned after motion onset.

This scenario has two levels of difficulty. In both, we place the target object
in the center of the grid marked with white tape in
Fig.~\ref{fig:setup_grasping_dynamic}.
Level 1: As soon as the gripper starts moving, we move the
target object to another grid cell. Level 2: We move and flip
the target object 180 degrees. For each level, this repositioning is
done very fast. For each level, we report execution time
and grasp success.

\begin{figure*}
\begin{subfigure}{.5\textwidth}
  \centering
  \includegraphics[width=0.8\columnwidth]{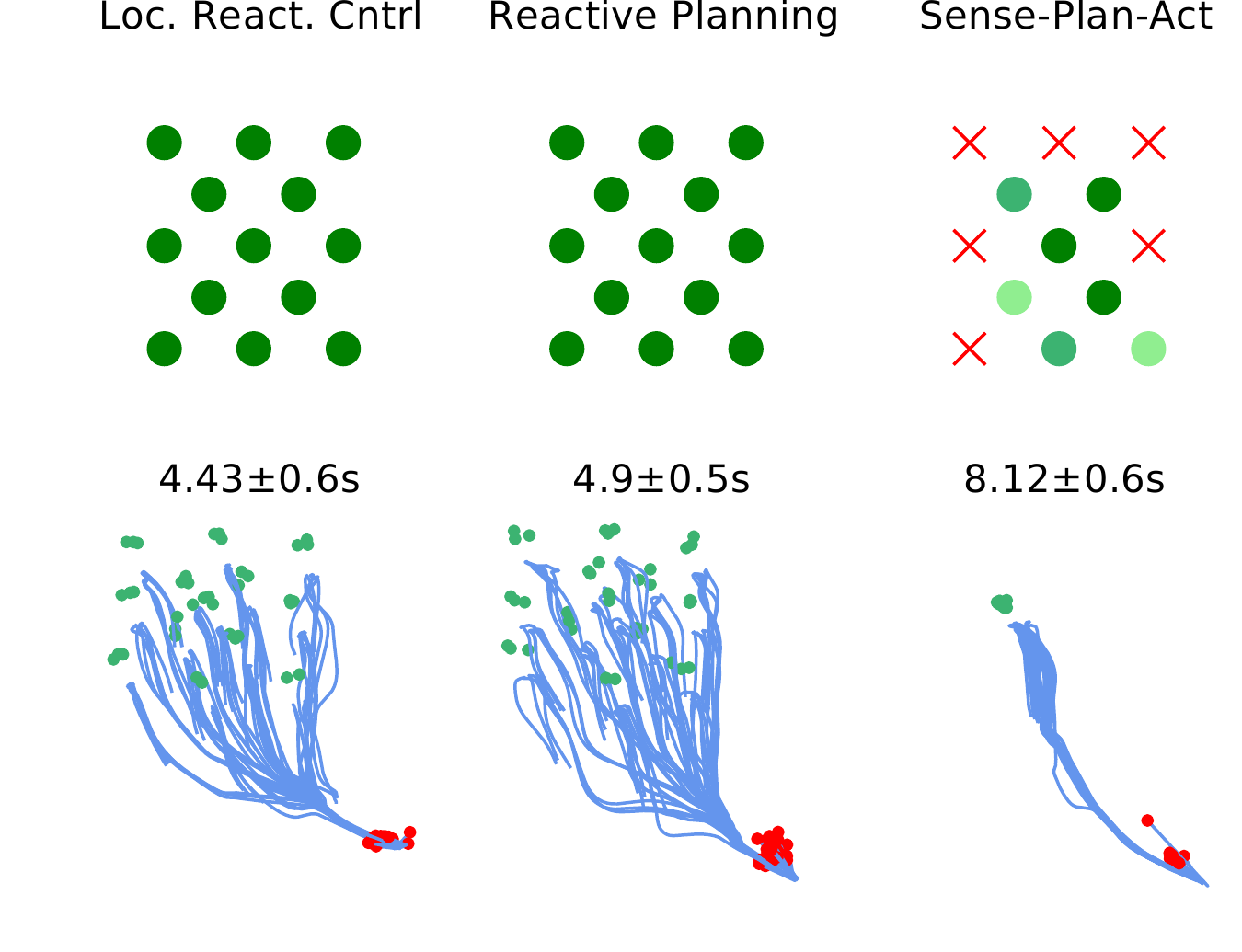}
  \caption{Target with Dynamic Position}
  \label{fig:target_motion} 
\end{subfigure}
\begin{subfigure}{.5\textwidth}
  \centering
  \includegraphics[width=0.8\columnwidth]{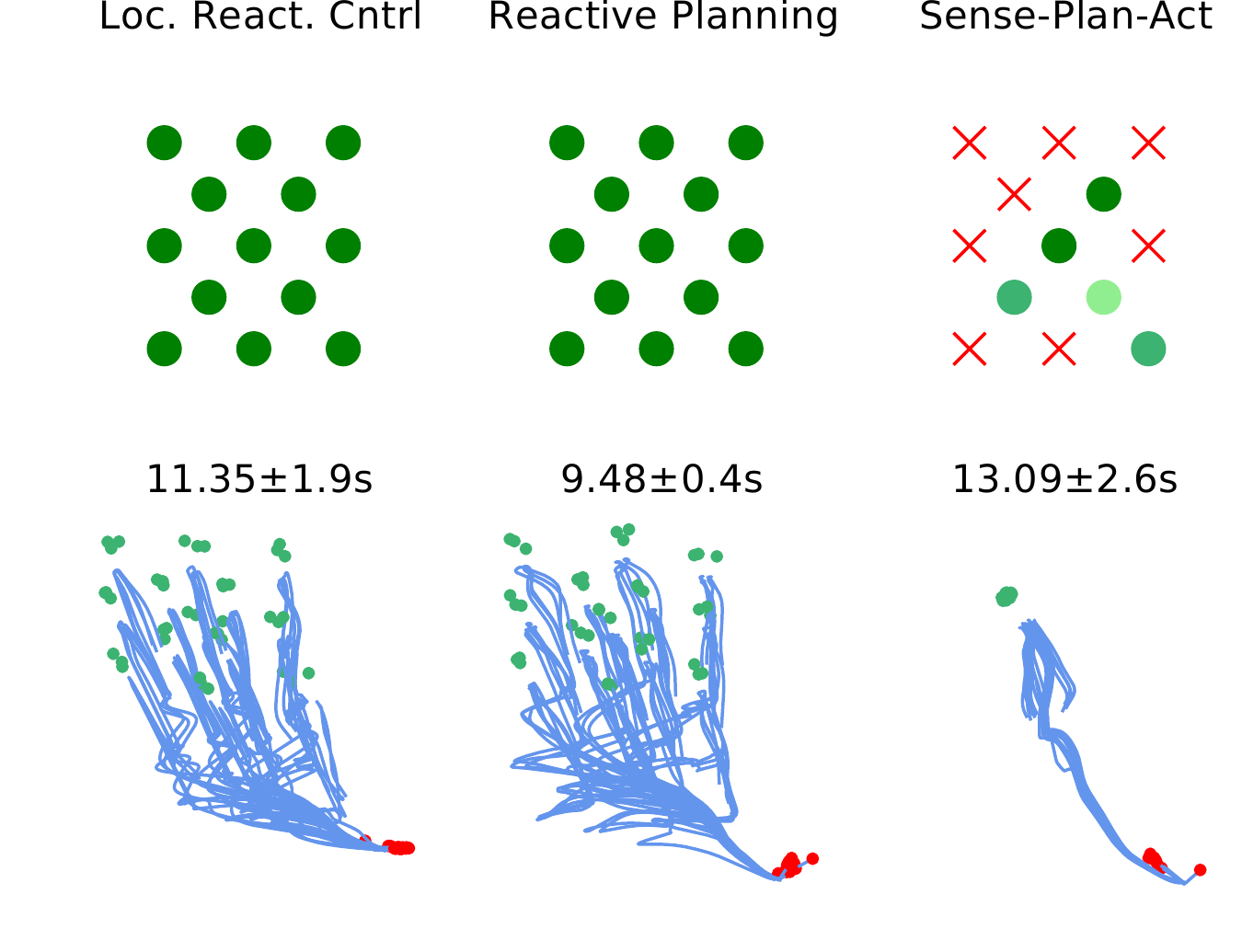}
  \caption{Target with Dynamic Pose}
  \label{fig:flipping_motion} 
\end{subfigure}
\caption{Visualization of successful grasps of in case of a (a)
  repositioned target and (b) repositioned and flipped target after
  motion onset. In both (a) and (b) we compare \lrc\ (Left), \rp\
  (Middle) and \spa\ (Right). (Top) Grid of possible target object
  poses after motion onset. The grid is $24\times 24cm$ large. The
  central dot marks the initial target object position. Dark green
  dots indicate that all three grasp attempts were successful. Red
  crosses indicate positions at which no grasp trial succeeded.
  (Bottom) Visualization of the end-effector trajectories in blue for
  successful grasps. Red dots indicate the starting position. Green
  dots indicate the target pose.(a) Both, \lrc\ and \rp\ grasp
  successfully in the entire region of variation and successfully
  adapt the trajectory given the new feedback on target pose. \Lrc\ is
  on average slightly faster than \rp\ (b) Again, both, \lrc\ and \rp\
  grasp successfully in the entire region of variation and
  successfully adapt the trajectory given the new feedback on target
  pose. Here, \rp\ is on average a bit faster. In both settings, \spa\
  manages to successfully grasp the target when the new position is
  in/close to the path to the original target location.}
\end{figure*}

For Level 1, Fig.~\ref{fig:target_motion} illustrates that both
\lrc\ and \rp\ are capable of adapting to the fast changing conditions
and successfully grasp the object for all possible changes. Notice,
\lrc\ is slightly faster in execution which can be explained by the
easy path planning problem required to solve this task. \Spa\ however
is only capable of grasping the object if it is close to the initial
position. For the more complex Level 2 shown in
Fig.~\ref{fig:flipping_motion}, we observe very similar performance
in terms of grasp success. Note, \spa\ did not adapt to the change of
orientation, the grasp just happened to work due to the symmetry of
the target object. In terms of execution speed, we observe a big difference
between \rp\ and \lrc\. The main reason for this
difference is that \rp\ continuously adapts to the rotated approach
whereas \lrc\ has an intrinsic tradeoff between convergence in
position and orientation. E.g in case of higher priority on
the position, \lrc\ will attempt to reduce the position error faster
compared to the orientation error. This can result in overall slower
convergence since orientation changes might be more difficult after
reaching the target position due to environmental constraints.
Arguably this tradeoff can be tuned to fit this scenario but it always
will be task specific whereas \rp\ automatically solves this
problem due the planning horizon.

This experiment emphasizes the importance of continuous feedback of
target object pose to adapt the grasping motion in case the object
moves during reaching. It also demonstrates the benefit of a longer planning horizon.

\subsubsection{Pointing in Dynamic Environments with Dynamic Target}
In our most complex scenario, we want to analyze the accuracy and
reactivity to simultaneous changes in the environment and target
pose. The task of the robot is to align its fingertip with the tip of
a drill (see Fig.~\ref{fig:setup_pointing_dynamic}).

We have four different levels of complexity for this task.
\textit{Level 1}: As a baseline we start with a static environment
without obstacles while the drill is stationary. \textit{Level 2}: we
introduce a blocking obstacle (box) during execution, which can be
avoided by going around it. \textit{Level 3}: the obstacle is moved
into the way such that the arm has to move over the obstacle or take a
big detour. \textit{Level 4}: We start out with a blocking obstacle.
After the system starts moving we remove the obstacle while also
changing the orientation of the drill by 90 degrees. The reorientation
of the drill means that the pointing approach has to be adapted.

\begin{table}
\centering
\begin{tabular}{l|c|c|c}%
  \bfseries Difficulty & \bfseries l. react. c. & \bfseries react. pl. & \bfseries s-p-a\\
  \hline 
  \hline 
  static   & 100\% (3)& 100\% (3)& 100\% (3)\\
  straight & 100\% (3)& 100\% (3)& 0\% (3)\\
  diagonal & 100\% (3)& 100\% (3)& 0\% (3)\\
  turning  & 100\% (3)& 100\% (3)& 0\% (3)\\
\end{tabular}
\caption{Success rates (total runs) of pointing experiment\label{tab:success_pointing}}
\end{table}
We report the results for this scenario in
Fig.~\ref{tab:success_pointing}. We define success as reaching the tip
of the drill up to a distance of 3 cm without a collision with any
environmental obstacle.
Neither \lrc\ nor \rp\ collide with any obstacle in this experiment.
\Spa\ however collides with the blocking obstacle for both the
\textit{Level 2} and \textit{Level 3} experiment. In the case of the
\textit{Level 4} experiment, \spa\ was able to reach the initial
position of the drill tip. Since no perceptual feedback is considered
it was not aware of the rotation whereas both \lrc\ and \rp\ could
even shorten the path towards the drill tip by taking into account the
removed obstacle.
For optimal performance, \textit{Level 4} requires both continuous tracking of the target
(similar to \textit{Grasping with Dynamic Targets}) and updates of the
workspace obstacles (similar to \textit{Pick and Place in Dynamic
  Environments}).

This experiment supports our hypothesis that integrating 
real-time perception with motion generation is key for
task success and  safe behavior in highly dynamic scenarios.

\section{Conclusion, Lessons Learned, Future Work}\label{sec:conclusion}
Already in the 80's~\cite{Brooks:1990}, it has been postulated that
tightly integrating real-time perception and reactive motion generation is
beneficial if not even required for robotic systems that
physically interact with uncertain and dynamic environments. 
To quantify the benefits of this integration for a
high DoF robotic manipulation system, we compared three different
systems (\rp, \lrc\ and \spa) with a varying level of integration
between perception and motion generation.

We have shown that already \lrc\ which integrates perceptual feedback
at the highest possible rate can be very efficient in simple tasks
while being safe by avoiding collisions with the environment. 
\RP\ achieves a better performance in more complex environments due to
its ability to look ahead. \Spa\ performs well in static scenarios as
expected, but even there \lrc\ and \RP\ have advantages as they can
start moving earlier while continuing to consume feedback.

We also observed that on the trade-off curve beween perceptual
accuracy and computational speed, it is more beneficial to have fast
feedback than accurate world representations. This is especially the
case for dynamic and uncertain manipulation scenarios where a fast
reaction to sudden changes or new incoming information is key. As
communication bandwidth is limited, this also places constraints and
how much information can be transferred between components. Therefore,
we opted for model-based visual tracking and querying SDFs only for a
small subset of points on the robot. Furthermore, we placed
computational nodes that interact frequently onto one computer
(vision, control, motion optimization). Data association was extremely
important and we therefore carefully synchronized the different
sensory and information streams across the three different computers. 
We observed that tuning parameters like safety margins
was straightforward as the underlying models of the system have an
intuitive interpretation. These parameters were also invariant across
the four different scenarios.

In the current architecture, we mostly take visual and joint encoder feedback
into account. Manipulation tasks are however heavily
concerned with contact interaction. We use the finger strain
gauges as feedback in the grasp controller.
Our system would however benefit from also taking haptic
feedback from tactile sensor arrays or force/torque sensors into
account~\cite{Kappler-RSS-15}. Currently, we are optimizing motion for
obstacle avoidance. However, exploiting contact constraints during
manipulation has been shown to increase
robustness~\cite{Righetti2014ARM,Kazemi2012ARM,isrr2013}.  
Our system also does not rely on any learning yet. However, there is
a large potential in e.g. learning representations of the perceptual
data instead of prescribing it ourselves. Another interesting research
direction is the integration of online inverse dynamics
learning to cope with changed dynamics after picking up
objects~\cite{DBLP:journals/corr/RatliffMKS16, franzi_iros_2016}. 

\section{Acknowledgments}
This research was supported in part by National Science Foundation
grants IIS-1205249, IIS-1017134, EECS-0926052, the Office of Naval
Research, the Okawa Foundation, and the Max-Planck-Society. Any
opinions, findings, and conclusions or recommendations expressed in
this material are those of the author(s) and do not necessarily
reflect the views of the funding organizations.

{\scriptsize
\bibliographystyle{IEEEtranS_short} 
\bibliography{references}
}
\end{document}